\documentclass[letterpaper]{article} 
\usepackage{aaai25}  
\nocopyright
\usepackage{times}  
\usepackage{helvet}  
\usepackage{courier}  
\usepackage[hyphens]{url}  
\usepackage{graphicx} 
\usepackage{natbib}  
\usepackage{caption} 
\usepackage{algorithm}
\usepackage{algorithmic}
\usepackage{amsfonts}
\usepackage{bm} 
\usepackage{amsmath} 
\usepackage{booktabs}
\usepackage{nicematrix} 
\usepackage{nicefrac} 
\usepackage{newfloat}
\usepackage{listings}
\newcommand{\iidsim}{\stackrel {\text{ { iid }}} {\sim} }

\DeclareCaptionStyle{ruled}{labelfont=normalfont,labelsep=colon,strut=off} 
\floatstyle{ruled}
\newfloat{listing}{tb}{lst}{}
\floatname{listing}{Listing}
\title{C3-Diff: Super-resolving Spatial Transcriptomics via Cross-modal Cross-content Contrastive Diffusion Modelling}
\author{
    Xiaofei Wang,\textsuperscript{\rm 1}
    Stephen Price,\textsuperscript{\rm 1}
    Chao Li,\textsuperscript{\rm 1,2,3,4}\thanks{Corresponding Author.}
    \\
}
\affiliations{
    \textsuperscript{\rm 1}Department of Clinical Neurosciences, University of Cambridge, UK\\
    \textsuperscript{\rm 2}Department of Applied Mathematics and Theoretical Physics, University of
 Cambridge, UK\\
    \textsuperscript{\rm 3}School of Science and Engineering, University of Dundee, UK\\
    \textsuperscript{\rm 4}School of Medicine, University of Dundee, UK\\

}
\usepackage{bibentry}

\begin{document}

\maketitle

\begin{abstract}
The rapid advancement of spatial transcriptomics (ST), i.e., spatial gene expressions, has made it possible to measure gene expression within original tissue, enabling us to discover molecular mechanisms.
However, current ST platforms frequently suffer from low resolution, limiting the in-depth understanding of spatial gene expression. Super-resolution approaches promise to enhance ST maps by integrating histology images with gene expressions of profiled tissue spots. However, it remains a challenge to model
the interactions between histology images and gene expressions for effective ST enhancement. This study presents a cross-modal cross-content contrastive diffusion framework, called C3-Diff, for ST enhancement with histology images as guidance. In C3-Diff, we firstly analyze the deficiency of traditional contrastive learning paradigm, which is then refined to extract both modal-invariant and content-invariant features of ST maps and histology images. 
Further, to overcome the problem of low sequencing sensitivity in ST maps, we perform  nosing-based information augmentation on the surface of feature unit hypersphere. Finally, we propose a dynamic cross-modal imputation-based training strategy to mitigate ST data scarcity. We tested C3-Diff  by benchmarking its performance on four public datasets, where it achieves significant improvements over competing methods. Moreover, we evaluate C3-Diff  on downstream  tasks of cell type localization, gene expression correlation and single-cell-level gene expression prediction, promoting AI-enhanced  biotechnology for biomedical research and clinical applications. Codes are available at \url{https://github.com/XiaofeiWang2018/C3-Diff}. 
\end{abstract}

\section{Introduction}
Gene expression captured by RNA sequencing offers in-depth insights into the molecular processes underlying biological systems. However, traditional RNA sequencing of bulk tissue only captures  overall expression patterns within a whole sample. As a further development, single-cell RNA sequencing (scRNA-seq) captures heterogeneity at the cellular resolution but still lacks spatial tissue context. Recently, spatial transcriptomics (ST), spatial distribution of gene expressions,  have emerged as a technique to profile the genomics of the tissue while preserving tissue structure, promising to characterise complex molecular processes inherently demonstrating spatial heterogeneity. 

Popular experimental ST methods, e.g., Visium \cite{visium} and SLIDE-seqV2 \cite{stickels2021highly}, only measure gene expression in tissue spots. The very low spatial resolution (e.g., 100 $\mu$m $\textrm{px}^{-1}$ of Visium) limits their ability to probe gene expression at cellular level (10 $\mu$m $\textrm{px}^{-1}$). Novel biotechnology is developed for high-resolution (HR) ST profiling, e.g., Xenium \cite{xenium}. However, these methods are expensive, time-consuming and  limited by the technical bottleneck of low capture  sensitivity.
 
Computational approaches promise to enhance the spatial resolution of ST maps \cite{zhang2024inferring}. Current approaches of enhancing ST maps  mainly  leverage paired scRNA-seq \cite{vahid2023high,longo2021integrating} providing gene expression of individual cells.  However, existing methods have achieved limited success \cite{he2024starfysh}, as they require paired single-cell data as reference, which is rather expensive and impractical \cite{he2024starfysh}.
On the other hand, high-resolution histology images \cite{hu2023deciphering} is enriched with cellular morphology features proven to be  associated with gene expression \cite{badea2020identifying}, which can provide crucial regional information compared to scRNA-seq data. As histology images are readily available for all ST maps, they could serve as an alternative for enhancing ST maps.  
However, cross-modal modeling of histology images and ST maps remains several challenges: 

Firstly, ST maps and histology images have shared and unique features crucial for biomedical research, i.e., histology images characterize phenotypic structure and cellular patterns, while ST maps bear unique features of expression patterns across genes. However, effective models to decode these features is still lacking.  
Secondly, Real-world technical limitations of ST, i.e., low profiling sensitivity, pose further challenge to effective modeling of ST data. Widely-used experimental ST techniques, e.g., Visium \cite{visium}, spatially barcode entire transcriptomes, but at limited capture rate of sequencing reads, causing  inevitable loss of expression value \cite{biancalani2021deep,rao2021exploring}.
Thirdly, Due to the  real-world scenarios of the ST data scarcity \cite{biancalani2021deep}, the histology images are often lack of paired reference of spot ST maps.

\begin{figure*}[t]
\includegraphics[width=.9\textwidth]{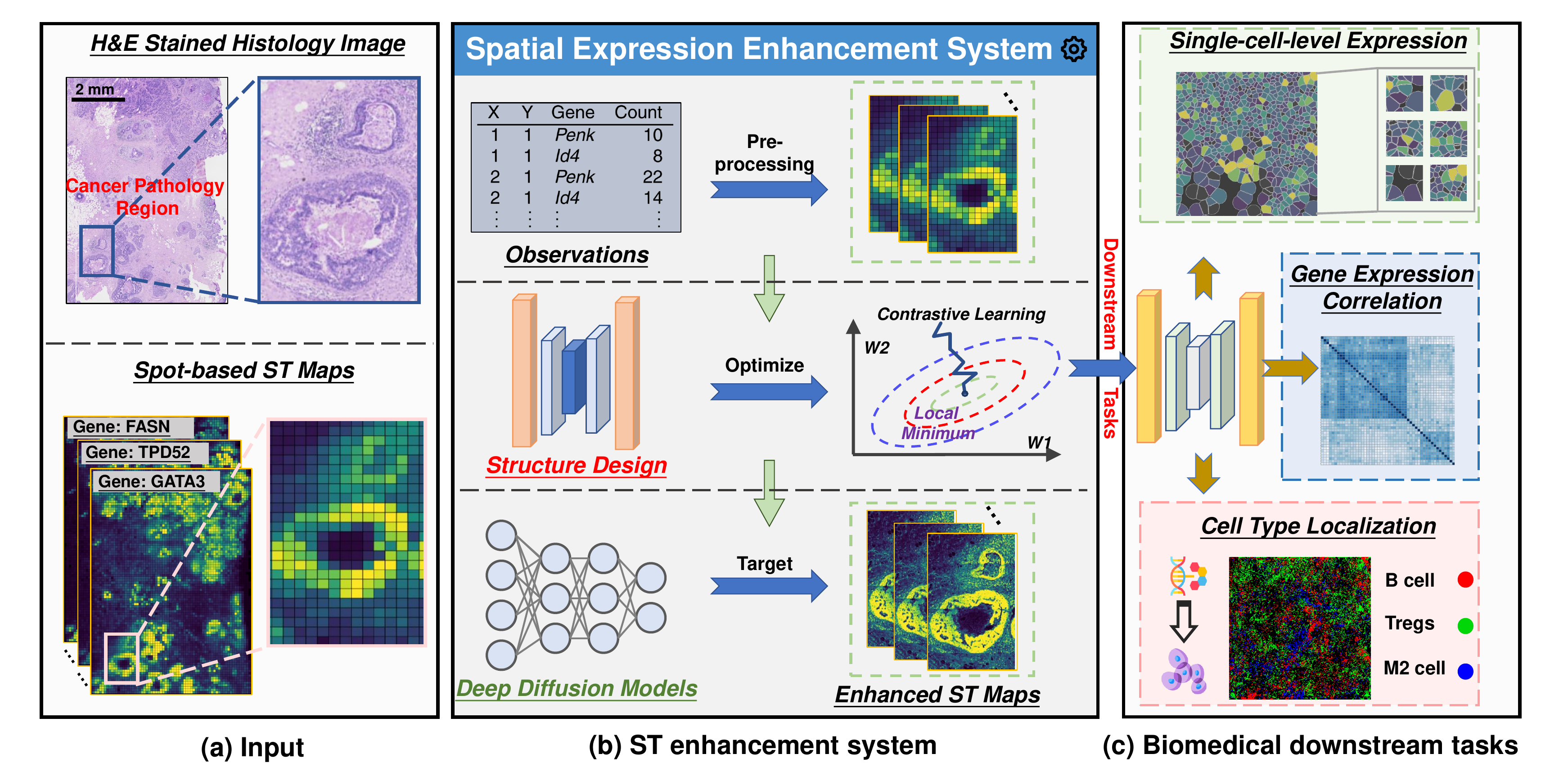}
\centering
\caption{ (a) Illustration of the  hematoxylin and eosin (H$\&$E) stained histology image and spot-based ST maps.  (b) Overview of the proposed ST enhancement system and (c) its downstream tasks.
 }
\label{Fig1}
\end{figure*}

This study presents C3-Diff (Cross-modal Cross-content Contrastive  Diffusion for ST Enhancement), a novel framework (Fig. \ref{Fig1}) to enhance spot ST maps based on histology images, inspired by the state-of-the-art (SOTA)  diffusion models in conditional image generation \cite{zhang2023adding,rombach2022high,zhao2024uni}. Detailed technical contributions are three-fold:
\textbf{\textcolor[rgb]{1.00,0.00,0.00}{(1)}} Despite success, these conditional diffusion models, such as Uni-controlnet \cite{zhao2024uni}, treat conditions equally without modelling interactions of multimodal conditions. To tackle this challenge, we design a novel cross-modal, cross-content contrastive learning method for bridging the translation from histology images to ST maps. Specifically, for efficient multimodal modelling, we firstly analyze the deficiency of traditional multimodal contrastive learning paradigm, which is then refined for extracting both modal-invariant and content-invariant features between histology images and ST maps. Of note, modal-invariant features indicate the inherent characteristics of a certain modality, e.g., cellular morphology in histology or expression patterns in ST maps. Meanwhile, content-invariant features indicate specific regions among patients informative of disease pathology, e.g., necrosis or microvascular proliferation. Then, we demonstrate the effectiveness of the modal-invariant and content-invariant features via the analysis of mutual information maximization.
\textbf{\textcolor[rgb]{1.00,0.00,0.00}{(2)}} To alleviate the limitation of low sensitivity in ST maps, we propose a noising-based information augmentation method on the surface of feature unit hypersphere, where Gaussian noise is injected to ST embeddings to mitigate information loss. 
\textbf{\textcolor[rgb]{1.00,0.00,0.00}{(3)}} We further propose a dynamic cross-modal omics imputation-based training strategy to tackle the data scarcity of ST.

We demonstrate the effectiveness of C3-Diff by benchmarking its performance using four public datasets of human breast and skin cancers, where it achieves significant improvements over both existing ST enhancement methods and SOTA conditional diffusion models.  Moreover, we further validate the biomedical impact of C3-Diff with predicted ST maps in three downstream tasks:  \noindent\textbf{i) Cell Type Localization}: generating the locations of different cell types in the tissue context. \noindent\textbf{ii) Gene Expression Correlation Analysis}: inferring the expression relationships across genes.\noindent\textbf{iii) Single-cell-level Expression Prediction}: predicting single-cell-level gene expression patterns.

To the best of our knowledge, this is the first cross-modal contrastive diffusion model for inferring enhanced  ST maps from histology images. The novel cross-modal cross-content contrastive diffusion framework is simple yet effective. The proposed framework promises to significantly reduce the cost associated with high-resolution gene profiling, promoting basic and clinical research on the avenue of  AI for  science in uncovering disease mechanisms and developing effective treatments.

\section{Related Work}

\subsection{Predicting ST maps from histology images}
Previous studies show that image-level histology features are associated with tissue gene expression patterns \cite{badea2020identifying,schmauch2020deep}. 
Therefore, several studies \cite{he2020integrating,xie2024spatially,jia2024thitogene} have made efforts in this direction,  e.g., \cite{he2020integrating} utilized ImageNet-pretrained DenseNet-121 \cite{huang2017densely} to successfully predict the spatial expression of 250 genes of breast cancer. Similarly,  \cite{xie2024spatially} proposed  a bi-modal contrastive-based framework (BLEEP) for predicting expression from histology images. However, these approaches only focus on predicting spot-based ST maps, thus incapable of enhancing the resolution of ST maps. 

To enhance the resolution of ST maps, some studies \cite{zhang2024inferring} have recently been proposed to super-resolve ST maps using histology images. For instance,  \cite{bergenstraahle2022super} proposed xFuse, a multi-scale latent generative model to enhance ST resolution via joint embeddings of histology features with spot ST maps. Similarly, \cite{zhang2024inferring} devised a Vision Transformers \cite{chen2022scaling}-based method, iStar, to infer HR ST maps. However, these  methods only use the spot ST as weak supervision, thus less capable of  modeling the cross-modal interactions between HR ST and histology images. 
Most recently, \cite{wang2024cross} proposed a diffusion-based model (Diff-ST) for ST enhancement. Nevertheless,  Diff-ST only focus on modal-invariant features without utilizing contrastive learning paradigm for efficient cross-modal modelling. Moreover, Diff-ST is incapable of enhancing ST when the paired spot ST is missing, greatly restricting its practical applications.
Different from these methods, we propose a C3-Diff framework for explicit integration of histology and ST maps to enhance ST resolution. 
Besides, our C3-Diff can predict HR ST when no LR ST map is available in testing, thus suitable for real-world scenarios.

\subsection{Conditional diffusion models}

Conditional diffusion models are a class of deep generative models that have achieved SOTA performance in natural and medical images \cite{croitoru2023diffusion,rombach2022high,zhang2023adding,zhao2024uni}. Generally, these models incorporate a Markov chain-based diffusion process for conditional image generation via specially designed conditioning mechanisms. For instance, Rombach \textit{et al.} \cite{rombach2022high} proposed latent diffusion models (LDM), where they augmented the underlying UNet \cite{ronneberger2015u} backbone with the cross-attention mechanism for the input conditional images. Despite effectiveness, LDM is designed for single modal condition and thus incapable for jointly learning multimodal conditions for ST enhancement. 

Recent efforts \cite{zhang2023adding,zhao2024uni} have been dedicated in introducing multimodal conditions into diffusion models. For instance, Zhang \textit{et al.} \cite{zhang2023adding} proposed ControlNet to
add spatial conditioning controls to large, pretrained text-to-image diffusion models. Similarly, Zhao \textit{et al.} \cite{zhao2024uni} devised a Uni-ControNet framework that allows for  smultaneously utilizing different local controls via a specially designed local control adapter. However, these  methods either simply adds (e.g., ControlNet) or concatenates (e.g., Uni-ControlNet)  multimodal features, without considering the shared and unique features of different modalities to achieve effective integration. In contrast, our method leverages cross-modal contrastive  learning in constructing the  conditioning mechanisms of diffusion models.

\subsection{Multimodal contrastive representation learning}

As an established self-supervised learning approach, contrastive learning \cite{wang2024multi,wang2020understanding} allows models to learn the knowledge behind data without explicit labels based on the InfoMax principle \cite{linsker1988self}. Generally, it aims to bring an anchor (i.e., data sample) closer to a similar instance and away from dissimilar instances, by optimizing their mutual information in the
embedding space. Recently, several  multimodal contrastive learning methods \cite{radford2021learning,mao2023contrastive,wang2023connecting} have been devised to encode different modalities into a semantically aligned shared space. For example, CLIP \cite{radford2021learning} and its variants \cite{sun2023eva,wang2023connecting} are proposed to align the shared features of paired texts and images.\textit{ However, we argue that traditional cross-modal contrastive learning  methods, including CLIP, mainly focus on aligning the semantics/content of the data from different modalities, thus less effective in extracting modality-specific features.} In contrast, the proposed C3-Diff can extract both modal-invariant and content-invariant features of histology images and expression maps, better facilitating the ST enhancement task.


\section{Methodology}

\subsection{Preliminaries}
\begin{figure*}[t]
\includegraphics[width=.8\textwidth]{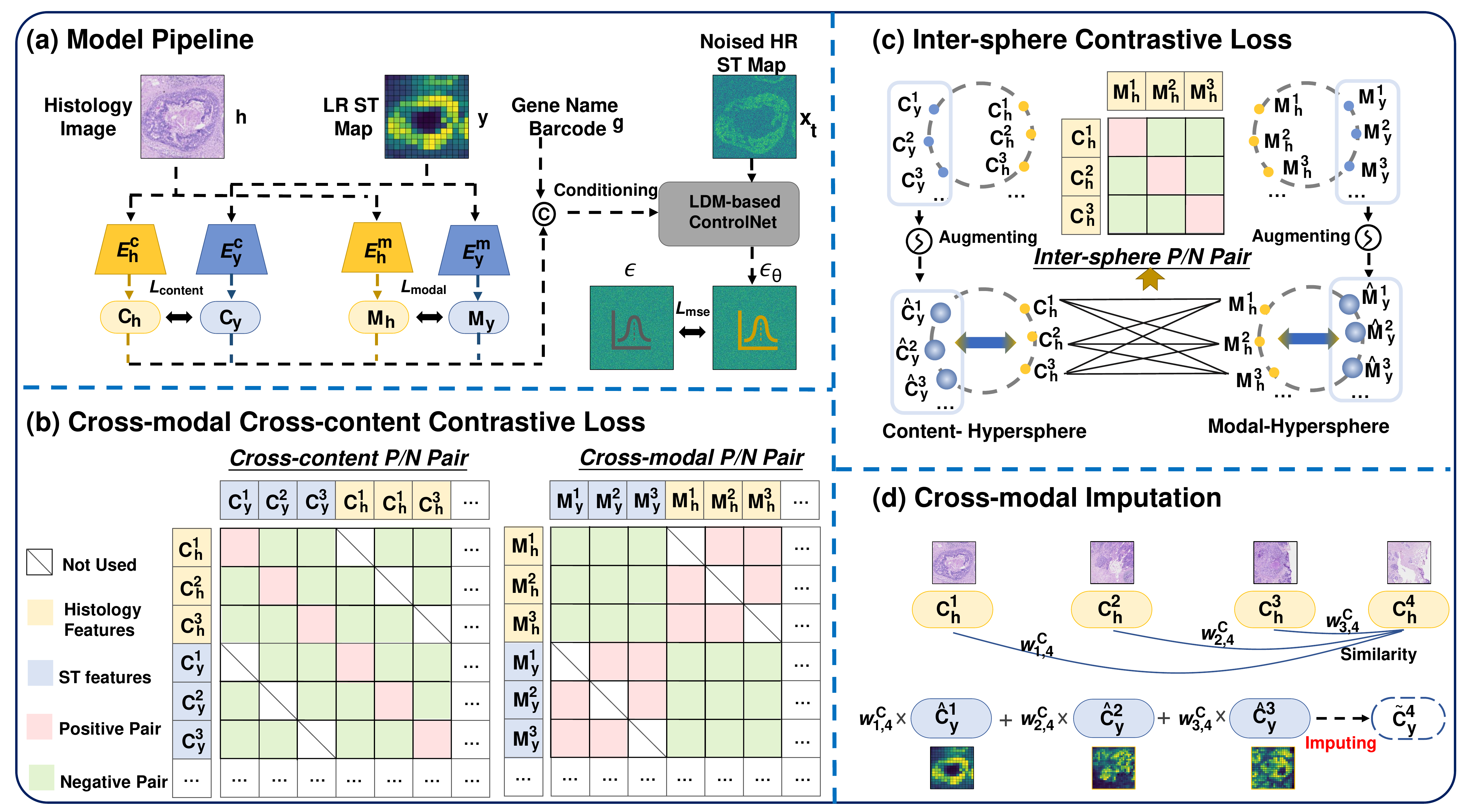}
\centering
\caption{ (a) Information flow of C3-Diff framework. (b) Positive/negative (P/N) pair construction for the cross-modal cross-content contrastive loss. (c) Illustration of information augmentation and inter-sphere loss.
 (d) Cross-modal imputation strategy for efficient learning with missing ST modality.
 }

\label{Fig2}
\end{figure*}

\noindent\textbf{Diffusion modelling.}
The proposed C3-Diff is inspired by a conditional diffusion model \cite{zhang2023adding}. As illustrated in Fig. \ref{Fig2}(a), our C3-Diff is trained to predict HR ST map $\mathbf{x}_0$ from Gaussian noise via an iterative denoising process,  conditioned on its paired LR ST map $\mathbf{y}$, histology image $\mathbf{h}$ and specific gene code $\mathbf{g}$.
The typical mean-squared error is used as the denoising objective:
\begin{align*}
  \mathcal{L}_{\rm mse} = \mathbb{E}_{\mathbf{x}_0, \mathbf{h}, \mathbf{y}, \bm{\epsilon}, t}(\|\bm{\epsilon} - \bm{\epsilon}_{\theta}(a_t \mathbf{x}_0 + \sigma_t \bm{\epsilon}, E(\mathbf{h}, \mathbf{y},\mathbf{g}))\|_2^2),
\end{align*}
where $E$ is the conditional feature generator, $t\sim \mathcal{U}(0, 1),~\bm{\epsilon}\sim \mathcal{N}(0, \mathbf{I})$ is the additive Gaussian noise, $a_t,\sigma_t$ are scalar functions of $t$, and $\bm{\epsilon}_{\theta}$ is a diffusion model with learnable parameters $\theta$. Besides, following \cite{rombach2022high,zhang2023adding}, \textit{Classifier-free guidance} is further employed for conditional data sampling, where the predicted noise is adjusted via:
\begin{align*}
  \hat{\bm{\epsilon}}_{\theta}(\mathbf{x}_t, E(\mathbf{h}, \mathbf{y},\mathbf{g})) = \omega \bm{\epsilon}_{\theta}(\mathbf{x}_t, E(\mathbf{h}, \mathbf{y},\mathbf{g})) + (1 - \omega) \bm{\epsilon}_{\theta}(\mathbf{x}_t),
\end{align*}
where $\mathbf{x}_t = a_t \mathbf{x}_0 + \sigma_t \bm{\epsilon}$, and $\omega$ is a guidance weight.
Detailed diffusion conditioning mechanism
is introduced as follows.

\noindent\textbf{Preliminaries on contrastive learning.}
The popular unsupervised contrastive representation learning method learns representations from unlabeled data. It assumes a way to sample \emph{positive pairs}, representing similar samples that should have similar representations. Empirically, the positive pairs are often obtained by taking two independently randomly augmented versions of the same sample \cite{chen2020simple}, or two samples of the same semantic content yet of different modalities \cite{radford2021learning}. Let $p_\mathrm{data}(\cdot)$ be the data distribution over $\mathbb{R}^n$ and $p_\mathrm{pos}(\cdot, \cdot)$ the distribution of positive pairs over $\mathbb{R}^n \times \mathbb{R}^n$. Then,  the contrastive loss \cite{he2020momentum}  can be formed as:

\begin{eqnarray*}
\mathcal{L}_{cl} = \mathcal{L}(z,z^+,z^-) = 
\mathbb{E}_{\substack{(z, z^+) \sim p_\mathrm{pos} \\ \{z^-\}_k \iidsim p_\mathrm{data}}} \\
\Big[- \log \big( \frac{\exp{(z^T \cdot z^+/\tau)}} {\exp{(z^T \cdot z^+/\tau)} + \sum_{k} \exp{(z^T \cdot z^-_k/\tau)}} \big)\Big],
\label{eq:contrastive_loss}
\end{eqnarray*}
where $\tau$ is a learnable temperature parameter. 

\noindent\textbf{Deficiency analysis of traditional contrastive learning.} 
One of the key parts of contrastive learning is positive/negative pairs construction, based on which the contrastive loss can be optimized.
The left table in Fig. \ref{Fig2}(b)  illustrates the positive and negative pair setting in a minibatch of three samples\footnote{For simplicity, we illustrate our contrastive learning settings with the minibatch of 3  in this paper.} of typical multimodal contrastive learning methods, e.g., CLIP \cite{radford2021learning}. In this table, $\mathbf{C}_{\mathrm{h}}^{i}$ represents the encoded features of histology image $\mathbf{h}_i$, and $\mathbf{C}_{\mathrm{y}}^{i}$ denotes the encoded features of the paired LR ST map $\mathbf{y}_i$. As seen, for feature $\mathbf{C}_{\mathrm{h}}^{1}$, $\mathbf{C}_{\mathrm{y}}^{1}$ forms its positive counterpart, since the two features represents the information from the same tissue sample.
 Meanwhile, $\{\mathbf{C}_{\mathrm{y}}^{2}, \mathbf{C}_{\mathrm{y}}^{3}, \mathbf{C}_{\mathrm{h}}^{2}, \mathbf{C}_{\mathrm{h}}^{3}\}$ denote the negative set, since they represents features of different tissue samples with that of $\mathbf{C}_{\mathrm{h}}^{1}$.

\textit{However, we argue that multimodal image features from the same  negative set should not be treated equally.} For instance, the negative pair $( \mathbf{C}_{\mathrm{h}}^{1}, \mathbf{C}_{\mathrm{h}}^{2})$ is from the same modality, while another negative pair $( \mathbf{C}_{\mathrm{h}}^{1}, \mathbf{C}_{\mathrm{y}}^{2})$ represents different modalities of histology images and ST.  Indeed, ST maps and histology images have modal-unique genetic and morphological information for ST enhancement.
Hence, the modality information should be further considered in constructing positive/negative pairs.

\subsection{Noising-based Information Augmentation on Unit Hypersphere.}
To generate the  features for contrastive learning, we propose a cross-modal feature extraction pipeline, illustrated in Fig. \ref{Fig2}(a). As shown in Fig. \ref{Fig2}(a), the input histology condition $\mathbf{h}$ is separately processed with the modality encoder $E_\mathrm{h}^\mathrm{m}$ and content encoder $E_\mathrm{h}^\mathrm{c}$, with the output of modal-related features $\mathbf{M}_\mathrm{h}$ and content-related features $\mathbf{C}_\mathrm{h}$, respectively. Similarly, $\mathbf{M}_\mathrm{y}$ and $\mathbf{C}_\mathrm{y}$ can be also generated 
for $\mathbf{y}$.
 
Besides, LR ST suffers from low sensitivity\footnote{Assuming average read per gene is 10,  then sensitivity level (error bar) of the gene expression value is $\pm 0.1$}, indicating that the ``real'' expression information could be lost.
When aligning existing representation spaces, this loss and bias of meaning will be inherited and amplified, affecting the robustness of alignment. To enhance the expression sensitivity of ST features, we propose to leverage Gaussian noise as an information augmentation method. Specifically, we add zero-mean Gaussian noises into ST features and re-normalize them to the unit hypersphere:

\begin{equation}\label{eqnoise}
\mathbf{\hat{M}}_\mathrm{y} = Norm(\mathbf{M}_\mathrm{y}+ \mu_1);    \;\; \mathbf{\hat{C}}_\mathrm{y} = Norm(\mathbf{C}_\mathrm{y}+ \mu_1)
\end{equation}

where noise items $\mu_1$ and $\mu_2$ are sampled from zero-mean gaussian distribution with variance $\sigma^2$.

\noindent\textbf{Augmenting mechanism.} As shown in Fig. \ref{Fig2}(c), each feature can be viewed as a point on the unit hypersphere, proven in \cite{wang2020understanding}.  The incorporation of Gaussian noise can transform the point into a small sphere, and re-normalizing projects the small sphere onto a circle of a new hypersphere. Features within the same circle share similar expressions, and the expression represented by the circle are more comprehensive and robust than the original point. This encourages the model to align embeddings of all the possible expressions of ST within those of histology images, thus alleviating the low sensitivity constraints. Moreover, when the expression of the target gene is not detected, the augmentation can still facilitate the robust representation learning by covering all possible situation including zero expressions.

\subsection{Cross-Modal Cross-Content Contrastive Learning}
Based on the generated features of different modalities (i.e., $\mathbf{M}_\mathrm{h}$ and $\mathbf{\hat{M}}_\mathrm{y}$) and content (i.e., $\mathbf{C}_\mathrm{h}$ and $\mathbf{\hat{C}}_\mathrm{y}$), we perform the cross-modal cross-content positive/negative (P/N) pair construction to extract and align modal-invariant and content-invariant features. The process of constructing P/N pairs is shown in Fig \ref{Fig2}(b). Specifically, based on the cross-modal P/N pairs, the cross-modal contrastive loss is defined as 

\begin{eqnarray*}\label{modalcl}
\mathcal{L}_{\text{modal}} = \mathbb{E}_{z\sim [\mathbf{M}_\mathrm{h}]_j, z^{+}\sim \mathcal{I}_{k\neq j}[\mathbf{M}_\mathrm{h}]_k, z^-\sim \mathcal{I}[\mathbf{\hat{M}}_\mathrm{y}]_k}  \mathcal{L}_{cl} \\
    + \mathbb{E}_{z\sim [\mathbf{\hat{M}}_\mathrm{y}]_j, z^{+}\sim \mathcal{I}_{k\neq j}[\mathbf{\hat{M}}_\mathrm{y}]_k, z^-\sim \mathcal{I}[\mathbf{M}_\mathrm{h}]_k}  \mathcal{L}_{cl} 
\end{eqnarray*}

Besides, the cross-content contrastive loss is based on the typical multimodal P/N setting \cite{radford2021learning}:

\begin{eqnarray*}\label{contentcl}
\mathcal{L}_{\text{content}} = \mathbb{E}_{z\sim [\mathbf{C}_\mathrm{h}]_j, z^{+}\sim [\mathbf{\hat{C}}_\mathrm{y}]_{j}, z^-\sim \mathcal{I}_{k\neq j}[\mathbf{C}_\mathrm{h},\mathbf{\hat{C}}_\mathrm{y}]_k}  \mathcal{L}_{cl} \\
    + \mathbb{E}_{z\sim [\mathbf{\hat{C}}_\mathrm{y}]_j, z^{+}\sim [\mathbf{C}_\mathrm{h}]_{j}, z^-\sim \mathcal{I}_{k\neq j}[\mathbf{\hat{C}}_\mathrm{y},\mathbf{C}_\mathrm{h}]_k}  \mathcal{L}_{cl}
\end{eqnarray*}

Moreover, according to \cite{wang2020understanding}, the optimization of the above two contrastive losses can be interpreted as the feature points alignment on the two respective unit hyperspheres, shown in Fig. \ref{Fig2}(c). Therefore, to better constrain the joint optimization of the embedded features of the two hyperspheres, we further propose an inter-sphere contrastive loss on features set $\{\mathbf{M}_\mathrm{h}, \mathbf{C}_\mathrm{h}\}$ as follows.

\begin{equation*}\label{intercl}
\mathcal{L}_{\text{inter-sphere}} = \mathbb{E}_{z\sim [\mathbf{M}_\mathrm{h}]_j, z^{+}\sim [\mathbf{C}_\mathrm{h}]_{j}, z^-\sim \mathcal{I}_{k\neq j}[\mathbf{C}_\mathrm{h}]_k}  \mathcal{L}_{cl} 
\end{equation*}

Of note, $\mathcal{L}_{\text{inter-sphere}}$ is similar to the contrastive loss in SimCLR \cite{chen2020simple}, where $\mathbf{C}_\mathrm{h}$ and $\mathbf{M}_\mathrm{h}$ can be regarded as the embedded features of two differently augmented versions of histology image $\mathbf{h}$.

\subsection{Dynamic Cross-modal Imputation-based Training Strategy }

Due to the data scarcity, the spot ST map could be missing, i.e., only histology images and gene names are available as the training conditions. To solve this modality-missing problem, we follow the idea of omics imputation \cite{song2020review} and propose a dynamic cross-modal imputation-based training strategy, as shown in Fig. \ref{Fig2}(d). The core idea is to impute the features of missing ST maps with those of the existing ST, weighted on the histology image-based correlation. Specifically,  given a minibatch of $N$ samples, where LR ST of $L$ samples is missing, the imputed ST features $\mathbf{\tilde{M}}_\mathrm{y}^l$ and $\mathbf{\tilde{C}}_\mathrm{y}^l$ of  $l$-th sample is  

\begin{eqnarray*}
\label{eqtrain}
\mathbf{\tilde{M}}_\mathrm{y}^l = \alpha \sum_{k=1}^{N-L} \underbrace{\frac{\mathrm{exp}(\mathbf{M}_\mathrm{h}^l \cdot\mathbf{M}_\mathrm{h}^k/\tau_1)}{\sum_{j=1}^{N-L} \mathrm{exp}(\mathbf{M}_\mathrm{h}^l \cdot\mathbf{M}_\mathrm{h}^j /\tau_1)}}_{w_{k,l}^{\mathbf{M}}} \ast \mathbf{\hat{M}}_\mathrm{y}^k;  \\
\mathbf{\tilde{C}}_\mathrm{y}^l = \beta\sum_{k=1}^{N-L} \underbrace{\frac{\mathrm{exp}(\mathbf{C}_\mathrm{h}^l \cdot\mathbf{C}_\mathrm{h}^k/\tau_1)}{\sum_{j=1}^{N-L} \mathrm{exp}(\mathbf{C}_\mathrm{h}^l \cdot\mathbf{C}_\mathrm{h}^j /\tau_1)}}_{w_{k,l}^{\mathbf{C}}} \ast \mathbf{\hat{C}}_\mathrm{y}^k,
\end{eqnarray*}

where $\tau_1$ is the temperature parameter, $\cdot$ denotes the operator for cosine distance, $w_{k,l}^{\mathbf{M}}$ and $w_{k,l}^{\mathbf{M}}$ are the imputation weight, $\alpha$ and $\beta$ are adjusting factors, which gradually decrease to zero in training, converting our strategy to the zero-padding method. The  zero-padding setting enables the model to predict HR ST maps without LR ST (i.e., only using gene names) even with batch size of one, which is common in practice. In addition, the overall loss for training the proposed diffusion model can be found in supplementary material.

\subsection{Mutual Information Maximization Analysis}
Here we demonstrate that the proposed $\mathcal{L}_{\text{modal}}$ and $\mathcal{L}_{\text{modal}}$ can help the model to learn modal-invariant and content-invariant features, respectively, via the analysis of mutual information maximization. 
Specifically, mutual information captures the nonlinear statistical dependencies between variables. For cross-modal contrastive loss $\mathcal{L}_{\text{modal}}$, the mutual information for the positive pair $(z,z^+) \sim ([\mathbf{M}_\mathrm{h}]_j,\mathcal{I}_{k\neq j}[\mathbf{M}_\mathrm{h}]_k)$ is defined as

\begin{eqnarray*}
    I(z,z^+)=\sum_{z,z^+}p(z,z^+)\log\frac{p(z,z^+)}{p(z)p(z^+)} \\
    =\sum_{z,z^+}p(z,z^+)\log\frac{p(z|z^+)}{p(z)}, 
\end{eqnarray*}

where $p(z|z^+)/p(z)$ represents the density ratio between $z$ and $z^+$. According to the proof in \cite{sugiyama2012density,sasaki2022representation}, the optimization of contrastive loss based on maximum likelihood estimation is equal to estimating the density ratio of positive training pairs. Therefore, with the optimization of $\mathcal{L}_{\text{modal}}$ , we can 
achieve mutual information maximization of the positive pair of $(z,z^+)$. Besides, the positive pair $([\mathbf{M}_\mathrm{h}]_j, \mathcal{I}_{k\neq j}[\mathbf{M}_\mathrm{h}]_k$ are from the same modality yet with different content, so that the $\mathcal{L}_{\text{modal}}$ can enable the model to learn modal-invariant features for ST enhancement.
More details about the demonstration of the content-invariant features can be seen in the supplementary material.

\section{Experiments}

\subsection{Datasets and Implementation Details}
Due to the page limit, the dataset preparation and implementation details\footnote{Code will be released upon acceptance} can be found in the supplementary material.

\begin{table*}[t!]
\scriptsize
\centering
\caption{Performance comparisons on three human breast cancer datasets with $5\times$ and $10\times$ enlargement scales. Bold numbers indicate the best results.}
\setlength{\tabcolsep}{.48em}{
\begin{tabular}{c||cc|cccc|cccc|cccc}
\toprule
\hline
Dataset     & \multicolumn{2}{c|}  {Attributes}     & \multicolumn{4}{c|}  {Breast-Xenium}  & \multicolumn{4}{c|}  {Breast-SGE} & \multicolumn{4}{c}  {Breast-ST}                                                    
\\ \hline
Scale         & \multicolumn{2}{c|}{}       & \multicolumn{2}{c|}{$5\times$}   & \multicolumn{2}{c|}{$10\times$}  & \multicolumn{2}{c|}{$5\times$}                               & \multicolumn{2}{c|}{$10\times$} & \multicolumn{2}{c|}{$5\times$}          & \multicolumn{2}{c}{$10\times$}                   
\\ \hline 

       & \multicolumn{1}{c}{For ST} & \multicolumn{1}{c|}{IMT**}   & \multicolumn{1}{c}{RMSE} & \multicolumn{1}{c|}{PCC} & \multicolumn{1}{c}{RMSE} & PCC & \multicolumn{1}{c}{RMSE} & \multicolumn{1}{c|}{PCC} & \multicolumn{1}{c}{RMSE} & PCC 
& \multicolumn{1}{c}{RMSE} & \multicolumn{1}{c|}{PCC} & \multicolumn{1}{c}{RMSE} & PCC\\ \hline \hline

U-Net     & \multicolumn{1}{c}{} & \multicolumn{1}{c|}{ }  & \multicolumn{1}{c}{0.385}    & \multicolumn{1}{c|}{0.178}     & \multicolumn{1}{c}{0.407}     & 0.192    & \multicolumn{1}{c}{0.356}     & \multicolumn{1}{c|}{0.484}     & \multicolumn{1}{c}{0.455}     &  0.545  & \multicolumn{1}{c}{0.328}     & \multicolumn{1}{c|}{0.519}     & \multicolumn{1}{c}{0.409}     & 0.528  \\

U-Net++   & \multicolumn{1}{c}{} & \multicolumn{1}{c|}{ } & \multicolumn{1}{c}{0.302}     & \multicolumn{1}{c|}{0.224}     & \multicolumn{1}{c}{0.289}     & 0.314    & \multicolumn{1}{c}{0.376}     & \multicolumn{1}{c|}{0.512}     & \multicolumn{1}{c}{0.434}     &  0.509  & \multicolumn{1}{c}{0.342}     & \multicolumn{1}{c|}{0.527}     & \multicolumn{1}{c}{0.325}     & 0.463  \\

AttenU-Net  & \multicolumn{1}{c}{} & \multicolumn{1}{c|}{ } & \multicolumn{1}{c}{0.385}     & \multicolumn{1}{c|}{0.162}     & \multicolumn{1}{c}{0.423}     & 0.196    & \multicolumn{1}{c}{0.337}     & \multicolumn{1}{c|}{0.402}     & \multicolumn{1}{c}{0.434}     &  0.367  & \multicolumn{1}{c}{0.326}     & \multicolumn{1}{c|}{0.501}     & \multicolumn{1}{c}{0.408}     & 0.434  \\ \hline

LDM  & \multicolumn{1}{c}{} & \multicolumn{1}{c|}{ }  & \multicolumn{1}{c}{0.317}     & \multicolumn{1}{c|}{0.286}     & \multicolumn{1}{c}{0.296}     & 0.331    & \multicolumn{1}{c}{0.315}     & \multicolumn{1}{c|}{0.493}     & \multicolumn{1}{c}{0.386}     &  0.493  & \multicolumn{1}{c}{0.236}     & \multicolumn{1}{c|}{0.578}     & \multicolumn{1}{c}{0.269}     & 0.576  \\ 

ControlNet  & \multicolumn{1}{c}{} & \multicolumn{1}{c|}{ }  & \multicolumn{1}{c}{0.219}     & \multicolumn{1}{c|}{0.315}     & \multicolumn{1}{c}{0.248}     & 0.324    & \multicolumn{1}{c}{0.286}     & \multicolumn{1}{c|}{0.547}     & \multicolumn{1}{c}{0.324}     &  0.509  & \multicolumn{1}{c}{0.186}     & \multicolumn{1}{c|}{0.627}     & \multicolumn{1}{c}{0.217}     & 0.648  \\

Uni-Control     & \multicolumn{1}{c}{} & \multicolumn{1}{c|}{\checkmark}    & \multicolumn{1}{c}{0.252}     & \multicolumn{1}{c|}{0.365}     & \multicolumn{1}{c}{0.240}     & 0.343    & \multicolumn{1}{c}{0.294}     & \multicolumn{1}{c|}{0.508}     & \multicolumn{1}{c}{0.339}     &  0.545  & \multicolumn{1}{c}{0.203}     & \multicolumn{1}{c|}{0.632}     & \multicolumn{1}{c}{0.209}     & 0.643  \\  \hline

HistoGene  & \multicolumn{1}{c}{\checkmark} & \multicolumn{1}{c|}{ }  & \multicolumn{1}{c}{0.235}     & \multicolumn{1}{c|}{0.262}     & \multicolumn{1}{c}{0.271}     & 0.328    & \multicolumn{1}{c}{0.315}     & \multicolumn{1}{c|}{0.501}     & \multicolumn{1}{c}{0.342}     &  0.508  & \multicolumn{1}{c}{0.243}     & \multicolumn{1}{c|}{0.606}     & \multicolumn{1}{c}{0.214}     & 0.580  \\

iStar      & \multicolumn{1}{c}{\checkmark} & \multicolumn{1}{c|}{ }    & \multicolumn{1}{c}{0.248}     & \multicolumn{1}{c|}{0.352}     & \multicolumn{1}{c}{0.247}     & 0.352    & \multicolumn{1}{c}{0.296}     & \multicolumn{1}{c|}{0.512}     & \multicolumn{1}{c}{0.338}     &  0.526  & \multicolumn{1}{c}{0.217}     & \multicolumn{1}{c|}{0.645}     & \multicolumn{1}{c}{0.213}     & 0.675  \\

TESLA   & \multicolumn{1}{c}{\checkmark} & \multicolumn{1}{c|}{ }    & \multicolumn{1}{c}{0.196}     & \multicolumn{1}{c|}{0.314}     & \multicolumn{1}{c}{0.235}     & 0.386    & \multicolumn{1}{c}{0.285}     & \multicolumn{1}{c|}{0.548}     & \multicolumn{1}{c}{0.312}     &  0.513  & \multicolumn{1}{c}{0.173}     & \multicolumn{1}{c|}{0.610}     & \multicolumn{1}{c}{0.207}     & 0.623  \\  

BLEEP   & \multicolumn{1}{c}{\checkmark} & \multicolumn{1}{c|}{ }    & \multicolumn{1}{c}{0.242}     & \multicolumn{1}{c|}{0.350}     & \multicolumn{1}{c}{0.245}     & 0.372    & \multicolumn{1}{c}{0.293}     & \multicolumn{1}{c|}{0.482}     & \multicolumn{1}{c}{0.296}     &  0.508  & \multicolumn{1}{c}{0.196}     & \multicolumn{1}{c|}{0.568}     & \multicolumn{1}{c}{0.244}     & 0.525  \\  

Diff-ST   & \multicolumn{1}{c}{\checkmark} & \multicolumn{1}{c|}{ }    & \multicolumn{1}{c}{0.168}     & \multicolumn{1}{c|}{0.346}     & \multicolumn{1}{c}{0.184}     & 0.392    & \multicolumn{1}{c}{0.224}     & \multicolumn{1}{c|}{0.542}     & \multicolumn{1}{c}{0.247}     &  0.525  & \multicolumn{1}{c}{0.175}     & \multicolumn{1}{c|}{0.622}     & \multicolumn{1}{c}{0.211}     & 0.618  \\  \hline

Ours (no LR ST)* & \multicolumn{1}{c}{\checkmark} & \multicolumn{1}{c|}{\checkmark} & \multicolumn{1}{c}{0.112}     & \multicolumn{1}{c|}{0.376}     & \multicolumn{1}{c}{0.148}     & 0.410     & \multicolumn{1}{c}{ 0.160}     & \multicolumn{1}{c|}{0.571}     & \multicolumn{1}{c}{0.196}    & 0.550 & \multicolumn{1}{c}{ 0.129}      & \multicolumn{1}{c|}{0.666}     & \multicolumn{1}{c}{0.140}     & 0.681   \\ 

\textbf{Ours} & \multicolumn{1}{c}{\checkmark} & \multicolumn{1}{c|}{\checkmark} & \multicolumn{1}{c}{\textbf{0.094}}     & \multicolumn{1}{c|}{\textbf{0.386}}     & \multicolumn{1}{c}{\textbf{0.137}}     & \textbf{0.432}     & \multicolumn{1}{c}{\textbf{0.146}}     & \multicolumn{1}{c|}{\textbf{0.582}}     & \multicolumn{1}{c}{\textbf{0.175}}     & \textbf{0.575} & \multicolumn{1}{c}{\textbf{0.103}}     & \multicolumn{1}{c|}{\textbf{0.693}}     & \multicolumn{1}{c}{\textbf{0.126}}     & \textbf{0.709}   \\ \hline
\bottomrule
\multicolumn{15}{l}{{ *\hspace{.3em} In model testing, the reference LR ST is replaced by zero padding maps. \; **\hspace{.3em} IMT refers to incomplete modality-based training. }}\\
\end{tabular}}
\label{tab1}

\end{table*}

\begin{figure}[t!]
    \centering
\includegraphics[width=.35\textwidth]{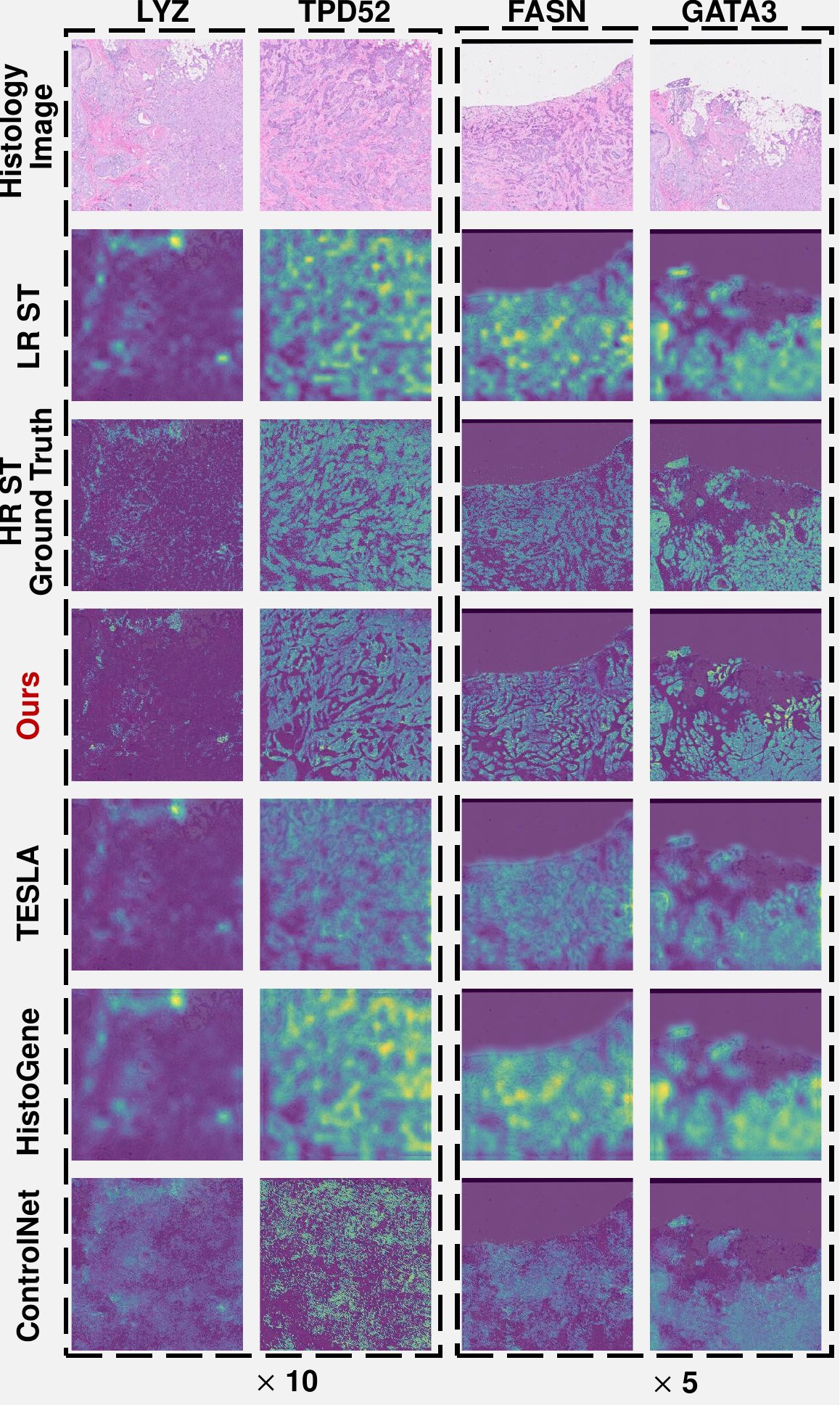}
    \caption{Visual comparisons at $5\times$ and $10\times$ scales on the Breast-Xenium dataset. 
    Note that GATA3, FASN, TPD52 and LYZ denote different genes.  }
    \label{fig:3}
    
\end{figure}

\subsection{Super-resolving Spatial Gene Expression}

We compare our model with ten other SOTA methods, i.e., iStar \cite{zhang2024inferring}, TESLA \cite{hu2023deciphering}, HistoGene \cite{pang2021leveraging}, BLEEP \cite{xie2024spatially}, Diff-ST \cite{wang2024cross}, LDM \cite{rombach2022high}, ControlNet \cite{zhang2023adding}, Uni-ControlNet \cite{zhao2024uni},  U-Net \cite{ronneberger2015u}, U-Net++ \cite{zhou2018unet++} and  AttenU-Net \cite{oktay1804attention}, at both 5$\times$ and 10$\times$ SR scales. Note iStar, TESLA, HistoGene, BLEEP and Diff-ST are specifically designed for ST SR tasks. Besides, LDM, ControlNet and Uni-ControlNet are SOTA conditional diffusion models, while other common image SR methods are baselines. 
To ensure fairness, all the comparison methods use both the HR histology image and LR ST maps for enhancing ST resolution. Moreover, the \textit{off-the-shelf representation} of the unconditional image synthesis task on CelebA-HQ dataset \cite{zhu2022celebv} is used for training the ControlNet.
As shown in Table \ref{tab1}, at 10$\times$ scale, C3-Diff performs the best, achieving improvement of at least 0.037 in Root MSE (RMSE) and 0.04 in Pearson correlation coefficient (PCC) over others, indicating that C3-Diff could successfully integrate histological features and gene expressions for ST SR. Similar results are also found in 5$\times$ scale.

In addition, we conduct experiments where no LR ST map is available in testing. As shown in Table \ref{tab1}, the results slightly decreases but still outperform all SOTA methods, indicating the potential of our method in real-world applications.
Moreover, Fig. \ref{fig:3} shows the subjective ST enhancement results of different methods at both $5\times$ and $10\times$ scales. C3-Diff outperforms all other methods, producing HR ST images with sharper edges and finer details. See more visual results in the supplementary material.

\begin{figure*}[t!]
    \centering
    \includegraphics[width=.8\textwidth]{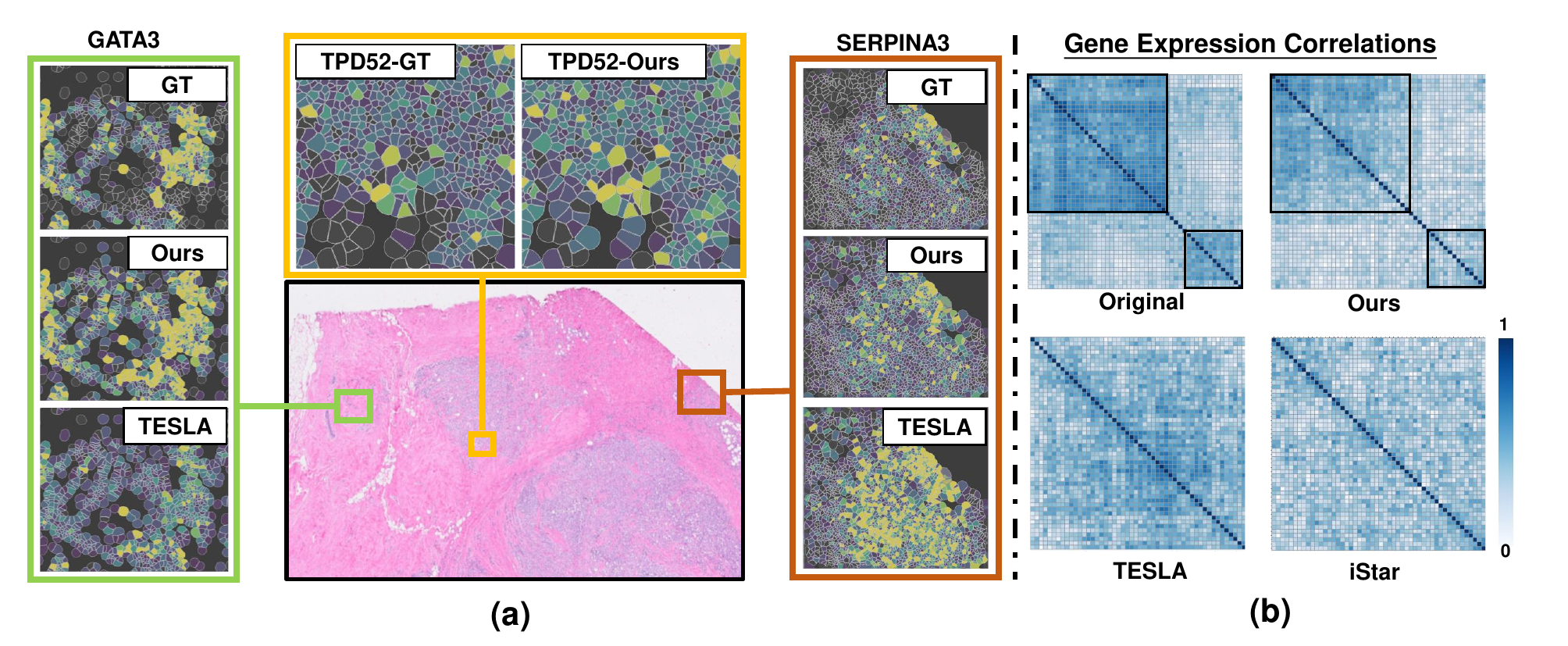}

    \caption{Results of the downstream tasks of (a) cell type localization and (b)  expression correlation analysis. GT denotes ground truth, while GATA3, TPD52 and SERPINA3 are  gene names.}
    \label{cell_expre}
    
\end{figure*}

\begin{minipage}{1\linewidth}
\scriptsize
\centering
\captionof{table}{Cross cancer validation on Melanoma. Performance comparisons on Melanoma-Xenium dataset with $10\times$ enlargement scales.}
\setlength{\tabcolsep}{2pt}
\resizebox{\linewidth}{!}{
\begin{NiceTabular}{c|c|c|c|c|c}
\toprule
 &U-Net          &U-Net++   &AttenU-Net     &LDM          &ControlNet \\
\hline
RMSE   &0.388 &0.292    &0.398   &0.334 & 0.276   \\
\hline
PCC   &0.217 &0.314    &0.226  &0.356 & 0.335   \\
\midrule \midrule
 
 &Uni-Control  &HistoGene      &iStar         &TESLA         &Ours  \\
\hline
RMSE  &0.227 &0.314    &0.242  &0.214  &\textbf{0.156}    \\\hline
PCC   &0.385 &0.318    &0.384   &0.395 & \textbf{0.478}   \\

\bottomrule
\end{NiceTabular}
}
\label{tab:5}
\end{minipage}%

\begin{minipage}{1\linewidth}
\scriptsize
\centering
\captionof{table}{Ablation Study on contrastive learning and our training strategy on the Breast-Xenium.}
\setlength{\tabcolsep}{2pt}
\resizebox{\linewidth}{!}{
\begin{tabular}{c||>{\centering\arraybackslash}p{1.2cm} >{\centering\arraybackslash} p{1.2cm}| >{\centering\arraybackslash} p{1.2cm} >{\centering\arraybackslash}p{1.2cm}}
\toprule
\hline
Scale            & \multicolumn{2}{c|}{5$\times$} & \multicolumn{2}{c}{10$\times$}    \\ \hline 
Metrics          & {RMSE} & {PCC} & {RMSE} & PCC \\ \hline \hline
$w/o$  augmentation    & 0.168   & {0.350}     &0.186    &  0.427  \\
$w/o$  $\mathcal{L}_{\mathrm{modal}}$       & {0.192}     &0.329    &  0.214 & 0.398  \\
$w/o$  $\mathcal{L}_{\mathrm{content}}$       & {0.188}     &0.324    &  0.202 & 0.414  \\
$w/o$  $\mathcal{L}_{\mathrm{inter-sphere}}$       & 0.145   & {0.326}     &0.178    &  0.420 \\ \hline \hline
Dropout                & {0.176}  &0.343        &0.192 & 0.401  \\
Zero padding           & 0.129   & {0.376}      &0.176    &  0.417  \\
Arithmetic average        & 0.114   & {0.356}     &0.167    &  0.406 \\ \hline \hline
\textbf{Ours}  & \textbf{0.094}     & {\textbf{0.386}}     & {\textbf{0.137}}     & \textbf{0.432}   \\ \hline
\bottomrule
\end{tabular}
}
\label{tab:6}
\end{minipage}%

\noindent\textbf{Cross Cancer Validation.} 
We further validate C3-Diff on skin cancers using the Melanoma-Xenium dataset. Results are shown
in Table \ref{tab:5} at the enlargement scale of 10. As shown, C3-Diff greatly outperforms all other SOTA methods by at least 0.058 in RMSE and 0.083 in PCC, indicating its effectiveness in ST enhancement on other cancers.

\noindent\textbf{Generalizability Validation.}  We further compare C3-Diff with other SOTA methods on two external validation datasets, i.e., Breast-SGE and Breast-ST, without fine-tuning, where both 5$\times$ and 10$\times$ SR scale settings are tested (Table \ref{tab1}). We observe that at 5$\times$ scale, our method achieves increments of 0.139 and 0.07 at RMSE, 
 and 0.034 and 0.048 at PCC, respectively,  compared to the best comparison method,suggesting the generalizability of C3-Diff.

\subsection{Downstream Task Validation}
We further evaluate our method on 3 downstream tasks:

\noindent\textbf{1) Gene Expression Correlation (GEC) Analysis}: 
GEC reveals the intrinsic correlation of co-expressed genes, suggesting genetic co-regulation mechanisms. We follow \cite{reynier2011importance} to generate the GEC by the predicted expressions of the involved 200 breast-caner related genes. Fig. \ref{cell_expre}(b) shows the comparison of C3-Diff and other SOTA methods. As shown, the generated GEC of C3-Diff better captures the detailed patterns of GEC, demonstrates the effectiveness of  C3-Diff in preserving gene-gene correlations and relevant biological heterogeneity.

\noindent\textbf{2) Single-cell-level Expression Prediction}: 
We further \textbf{quantitatively} assess C3-Diff’s ability to predict single-cell-level gene expression (Fig. \ref{cell_expre}(a)). Specifically, the predicted single-cell-level gene expression is computed from the super-resolved expressions using the cell segmentation masks provided in \cite{xenium_brest}. As shown, C3-Diff can better predict single-cell gene expression than other SOTA methods. Note that the genes in Fig. \ref{cell_expre}(a), i.e., GATA3, TPD52 and SERPINA3, are all key genes in breast cancer, indicating the potential of C3-Diff in downstream cellular level discovery and precision oncology.

\noindent\textbf{3) Cell Type Localization}: 
Due to the page limit, this part can be found in the supplementary material.

\subsection{Results of Ablation Experiments}

\noindent\textbf{1) Ablation on Contrastive Learning}
We assess the the proposed cross-modal cross-content contrastive learning method as follows: 
1) $w/o$ information augmentation - utilize the original embedded ST features; 2) $w/o$ cross-modal contrastive loss - remove $\mathcal{L}_{\text{modal}}$; 3) $w/o$ cross-content contrastive loss - remove $\mathcal{L}_{\text{cont}}$; 4) $w/o$ inter-sphere contrastive loss - remove $\mathcal{L}_{\text{inter-sphere}}$. The results on Breast-Xenium dataset are in Table \ref{tab:6}. All three models perform worse than C3-Diff, suggesting that these components can enhance the overall model performance. 
Moreover, $w/o$ $\mathcal{L}_{\text{modal}}$ performs the worst, consistent with our hypothesis that traditional contrastive loss may not effectively leverage modality-related information.

\noindent\textbf{2) Ablation on Cross-modal Imputation-based Training Strategy} 
We replace our training strategy with other three schemes: 
1) dropout - remove the all modality-incomplete training samples; 2) zero padding - replace the missing ST with zero maps; 3) arithmetic average  - replace the weight average to arithmetic average. Table \ref{tab:6} shows the results,   where all 3 modality-missing training methods perform worse than C3-Diff, indicating the effectiveness of our cross-modal imputation-based training strategy.

\section{ Discussion and Conclusion}

ST is an advanced biotechnology but is restricted by low spatial resolution for in-depth biomedical research. We propose C3-Diff, a novel framework based on conditional diffusion model for ST enhancement. We devise a cross-modal cross-content contrastive learning method to extract modal-invariant and content-invariant features to model interaction of histology images and ST maps.
To mitigate the limitation of the low  sensitivity of ST maps, we propose an information augmentation method to robustly align the ST and histology features.
In addition, a dynamic cross-modal imputation-based training strategy is  designed to  alleviate the real-world restriction of ST data scarcity.
Our experiments demonstrate that C3-Diff achieves superior and more robust performance over other state-of-the-art methods, in  spatial gene expression enhancement and various downstream tasks, opening a new avenue of AI-enhancing ST for  biomedical research and clinical application.
Our main limitation lies in the number of predicted genes, i.e., this study only predicts the expression of 200 marker genes in breast and skin cancer. Future work could improve the model by involving  whole transcriptomes, also exploring the inherent correlation across genes.

\bibliography{aaai25}

\begin{thebibliography}{42}
\providecommand{\natexlab}[1]{#1}

\bibitem[{Badea and St{\u{a}}nescu(2020)}]{badea2020identifying}
Badea, L.; and St{\u{a}}nescu, E. 2020.
\newblock Identifying transcriptomic correlates of histology using deep
  learning.
\newblock \emph{PloS one}, 15(11): e0242858.

\bibitem[{Bergenstr{\aa}hle et~al.(2022)Bergenstr{\aa}hle, He,
  Bergenstr{\aa}hle, Abalo, Mirzazadeh, Thrane, Ji, Andersson, Larsson,
  Stakenborg et~al.}]{bergenstraahle2022super}
Bergenstr{\aa}hle, L.; He, B.; Bergenstr{\aa}hle, J.; Abalo, X.; Mirzazadeh,
  R.; Thrane, K.; Ji, A.~L.; Andersson, A.; Larsson, L.; Stakenborg, N.; et~al.
  2022.
\newblock Super-resolved spatial transcriptomics by deep data fusion.
\newblock \emph{Nature biotechnology}, 40(4): 476--479.

\bibitem[{Biancalani et~al.(2021)Biancalani, Scalia, Buffoni, Avasthi, Lu,
  Sanger, Tokcan, Vanderburg, Segerstolpe, Zhang et~al.}]{biancalani2021deep}
Biancalani, T.; Scalia, G.; Buffoni, L.; Avasthi, R.; Lu, Z.; Sanger, A.;
  Tokcan, N.; Vanderburg, C.~R.; Segerstolpe, {\AA}.; Zhang, M.; et~al. 2021.
\newblock Deep learning and alignment of spatially resolved single-cell
  transcriptomes with Tangram.
\newblock \emph{Nature methods}, 18(11): 1352--1362.

\bibitem[{Chen et~al.(2022)Chen, Chen, Li, Chen, Trister, Krishnan, and
  Mahmood}]{chen2022scaling}
Chen, R.~J.; Chen, C.; Li, Y.; Chen, T.~Y.; Trister, A.~D.; Krishnan, R.~G.;
  and Mahmood, F. 2022.
\newblock Scaling vision transformers to gigapixel images via hierarchical
  self-supervised learning.
\newblock In \emph{Proceedings of the IEEE/CVF Conference on Computer Vision
  and Pattern Recognition}, 16144--16155.

\bibitem[{Chen et~al.(2020)Chen, Kornblith, Norouzi, and
  Hinton}]{chen2020simple}
Chen, T.; Kornblith, S.; Norouzi, M.; and Hinton, G. 2020.
\newblock A simple framework for contrastive learning of visual
  representations.
\newblock In \emph{International conference on machine learning}, 1597--1607.
  PMLR.

\bibitem[{Croitoru et~al.(2023)Croitoru, Hondru, Ionescu, and
  Shah}]{croitoru2023diffusion}
Croitoru, F.-A.; Hondru, V.; Ionescu, R.~T.; and Shah, M. 2023.
\newblock Diffusion models in vision: A survey.
\newblock \emph{IEEE Transactions on Pattern Analysis and Machine
  Intelligence}.

\bibitem[{Du et~al.(2024)Du, Wang, Law, Amann-Zalcenstein, Anttila, Ling,
  Hickey, Sargeant, Chen, Ioannidis et~al.}]{visium}
Du, M.~R.; Wang, C.; Law, C.~W.; Amann-Zalcenstein, D.; Anttila, C.~J.; Ling,
  L.; Hickey, P.~F.; Sargeant, C.~J.; Chen, Y.; Ioannidis, L.~J.; et~al. 2024.
\newblock Spotlight on 10x Visium: a multi-sample protocol comparison of
  spatial technologies.
\newblock \emph{bioRxiv}, 2024--03.

\bibitem[{He et~al.(2020{\natexlab{a}})He, Bergenstr{\aa}hle, Stenbeck, Abid,
  Andersson, Borg, Maaskola, Lundeberg, and Zou}]{he2020integrating}
He, B.; Bergenstr{\aa}hle, L.; Stenbeck, L.; Abid, A.; Andersson, A.; Borg,
  {\AA}.; Maaskola, J.; Lundeberg, J.; and Zou, J. 2020{\natexlab{a}}.
\newblock Integrating spatial gene expression and breast tumour morphology via
  deep learning.
\newblock \emph{Nature biomedical engineering}, 4(8): 827--834.

\bibitem[{He et~al.(2020{\natexlab{b}})He, Fan, Wu, Xie, and
  Girshick}]{he2020momentum}
He, K.; Fan, H.; Wu, Y.; Xie, S.; and Girshick, R. 2020{\natexlab{b}}.
\newblock Momentum contrast for unsupervised visual representation learning.
\newblock In \emph{Proceedings of the IEEE/CVF conference on computer vision
  and pattern recognition}, 9729--9738.

\bibitem[{He et~al.(2024)He, Jin, Nazaret, Shi, Chen, Rampersaud, Dhillon,
  Valdez, Friend, Fan et~al.}]{he2024starfysh}
He, S.; Jin, Y.; Nazaret, A.; Shi, L.; Chen, X.; Rampersaud, S.; Dhillon,
  B.~S.; Valdez, I.; Friend, L.~E.; Fan, J.~L.; et~al. 2024.
\newblock Starfysh integrates spatial transcriptomic and histologic data to
  reveal heterogeneous tumor--immune hubs.
\newblock \emph{Nature Biotechnology}, 1--13.

\bibitem[{Hu et~al.(2023)Hu, Coleman, Zhang, Lee, Kadara, Wang, and
  Li}]{hu2023deciphering}
Hu, J.; Coleman, K.; Zhang, D.; Lee, E.~B.; Kadara, H.; Wang, L.; and Li, M.
  2023.
\newblock Deciphering tumor ecosystems at super resolution from spatial
  transcriptomics with TESLA.
\newblock \emph{Cell systems}, 14(5): 404--417.

\bibitem[{Huang et~al.(2017)Huang, Liu, Van Der~Maaten, and
  Weinberger}]{huang2017densely}
Huang, G.; Liu, Z.; Van Der~Maaten, L.; and Weinberger, K.~Q. 2017.
\newblock Densely connected convolutional networks.
\newblock In \emph{Proceedings of the IEEE conference on computer vision and
  pattern recognition}, 4700--4708.

\bibitem[{Janesick et~al.(2023)Janesick, Shelansky, Gottscho, Wagner, Williams,
  Rouault, Beliakoff, Morrison, Oliveira, Sicherman et~al.}]{xenium_brest}
Janesick, A.; Shelansky, R.; Gottscho, A.~D.; Wagner, F.; Williams, S.~R.;
  Rouault, M.; Beliakoff, G.; Morrison, C.~A.; Oliveira, M.~F.; Sicherman,
  J.~T.; et~al. 2023.
\newblock High resolution mapping of the tumor microenvironment using
  integrated single-cell, spatial and in situ analysis.
\newblock \emph{Nature Communications}, 14(1): 8353.

\bibitem[{Jia et~al.(2024)Jia, Liu, Chen, Zhao, and Wang}]{jia2024thitogene}
Jia, Y.; Liu, J.; Chen, L.; Zhao, T.; and Wang, Y. 2024.
\newblock THItoGene: a deep learning method for predicting spatial
  transcriptomics from histological images.
\newblock \emph{Briefings in Bioinformatics}, 25(1): bbad464.

\bibitem[{Linsker(1988)}]{linsker1988self}
Linsker, R. 1988.
\newblock Self-organization in a perceptual network.
\newblock \emph{Computer}, 21(3): 105--117.

\bibitem[{Longo et~al.(2021)Longo, Guo, Ji, and Khavari}]{longo2021integrating}
Longo, S.~K.; Guo, M.~G.; Ji, A.~L.; and Khavari, P.~A. 2021.
\newblock Integrating single-cell and spatial transcriptomics to elucidate
  intercellular tissue dynamics.
\newblock \emph{Nature Reviews Genetics}, 22(10): 627--644.

\bibitem[{Mao et~al.(2023)Mao, Zhang, Xiang, Lv, Zhong, and
  Dai}]{mao2023contrastive}
Mao, Y.; Zhang, J.; Xiang, M.; Lv, Y.; Zhong, Y.; and Dai, Y. 2023.
\newblock Contrastive conditional latent diffusion for audio-visual
  segmentation.
\newblock \emph{arXiv preprint arXiv:2307.16579}.

\bibitem[{Oktay et~al.(1804)Oktay, Schlemper, Folgoc, Lee, Heinrich, Misawa,
  Mori, McDonagh, Hammerla, Kainz et~al.}]{oktay1804attention}
Oktay, O.; Schlemper, J.; Folgoc, L.~L.; Lee, M.; Heinrich, M.; Misawa, K.;
  Mori, K.; McDonagh, S.; Hammerla, N.~Y.; Kainz, B.; et~al. 1804.
\newblock Attention u-net: Learning where to look for the pancreas. arXiv 2018.
\newblock \emph{arXiv preprint arXiv:1804.03999}.

\bibitem[{Pang, Su, and Li(2021)}]{pang2021leveraging}
Pang, M.; Su, K.; and Li, M. 2021.
\newblock Leveraging information in spatial transcriptomics to predict
  super-resolution gene expression from histology images in tumors.
\newblock \emph{BioRxiv}, 2021--11.

\bibitem[{Radford et~al.(2021)Radford, Kim, Hallacy, Ramesh, Goh, Agarwal,
  Sastry, Askell, Mishkin, Clark et~al.}]{radford2021learning}
Radford, A.; Kim, J.~W.; Hallacy, C.; Ramesh, A.; Goh, G.; Agarwal, S.; Sastry,
  G.; Askell, A.; Mishkin, P.; Clark, J.; et~al. 2021.
\newblock Learning transferable visual models from natural language
  supervision.
\newblock In \emph{International conference on machine learning}, 8748--8763.
  PMLR.

\bibitem[{Rao et~al.(2021)Rao, Barkley, Fran{\c{c}}a, and
  Yanai}]{rao2021exploring}
Rao, A.; Barkley, D.; Fran{\c{c}}a, G.~S.; and Yanai, I. 2021.
\newblock Exploring tissue architecture using spatial transcriptomics.
\newblock \emph{Nature}, 596(7871): 211--220.

\bibitem[{Reynier et~al.(2011)Reynier, Petit, Paye, Turrel-Davin, Imbert, Hot,
  Mougin, and Miossec}]{reynier2011importance}
Reynier, F.; Petit, F.; Paye, M.; Turrel-Davin, F.; Imbert, P.-E.; Hot, A.;
  Mougin, B.; and Miossec, P. 2011.
\newblock Importance of correlation between gene expression levels: application
  to the type I interferon signature in rheumatoid arthritis.
\newblock \emph{PloS one}, 6(10): e24828.

\bibitem[{Rombach et~al.(2022)Rombach, Blattmann, Lorenz, Esser, and
  Ommer}]{rombach2022high}
Rombach, R.; Blattmann, A.; Lorenz, D.; Esser, P.; and Ommer, B. 2022.
\newblock High-resolution image synthesis with latent diffusion models.
\newblock In \emph{Proceedings of the IEEE/CVF conference on computer vision
  and pattern recognition}, 10684--10695.

\bibitem[{Ronneberger, Fischer, and Brox(2015)}]{ronneberger2015u}
Ronneberger, O.; Fischer, P.; and Brox, T. 2015.
\newblock U-net: Convolutional networks for biomedical image segmentation.
\newblock In \emph{Medical Image Computing and Computer-Assisted
  Intervention--MICCAI 2015: 18th International Conference, Munich, Germany,
  October 5-9, 2015, Proceedings, Part III 18}, 234--241. Springer.

\bibitem[{Salas et~al.(2023)Salas, Czarnewski, Kuemmerle, Helgadottir,
  Matsson-Langseth, Tismeyer, Avenel, Rehman, Tiklova, Andersson
  et~al.}]{xenium}
Salas, S.~M.; Czarnewski, P.; Kuemmerle, L.~B.; Helgadottir, S.;
  Matsson-Langseth, C.; Tismeyer, S.; Avenel, C.; Rehman, H.; Tiklova, K.;
  Andersson, A.; et~al. 2023.
\newblock Optimizing Xenium In Situ data utility by quality assessment and best
  practice analysis workflows.
\newblock \emph{BioRxiv}, 2023--02.

\bibitem[{Sasaki and Takenouchi(2022)}]{sasaki2022representation}
Sasaki, H.; and Takenouchi, T. 2022.
\newblock Representation learning for maximization of MI, nonlinear ICA and
  nonlinear subspaces with robust density ratio estimation.
\newblock \emph{Journal of Machine Learning Research}, 23(231): 1--55.

\bibitem[{Schmauch et~al.(2020)Schmauch, Romagnoni, Pronier, Saillard,
  Maill{\'e}, Calderaro, Kamoun, Sefta, Toldo, Zaslavskiy
  et~al.}]{schmauch2020deep}
Schmauch, B.; Romagnoni, A.; Pronier, E.; Saillard, C.; Maill{\'e}, P.;
  Calderaro, J.; Kamoun, A.; Sefta, M.; Toldo, S.; Zaslavskiy, M.; et~al. 2020.
\newblock A deep learning model to predict RNA-Seq expression of tumours from
  whole slide images.
\newblock \emph{Nature communications}, 11(1): 3877.

\bibitem[{Song et~al.(2020)Song, Greenbaum, Luttrell~IV, Zhou, Wu, Shen, Gong,
  Zhang, and Deng}]{song2020review}
Song, M.; Greenbaum, J.; Luttrell~IV, J.; Zhou, W.; Wu, C.; Shen, H.; Gong, P.;
  Zhang, C.; and Deng, H.-W. 2020.
\newblock A review of integrative imputation for multi-omics datasets.
\newblock \emph{Frontiers in Genetics}, 11: 570255.

\bibitem[{Stickels et~al.(2021)Stickels, Murray, Kumar, Li, Marshall, Di~Bella,
  Arlotta, Macosko, and Chen}]{stickels2021highly}
Stickels, R.~R.; Murray, E.; Kumar, P.; Li, J.; Marshall, J.~L.; Di~Bella,
  D.~J.; Arlotta, P.; Macosko, E.~Z.; and Chen, F. 2021.
\newblock Highly sensitive spatial transcriptomics at near-cellular resolution
  with Slide-seqV2.
\newblock \emph{Nature biotechnology}, 39(3): 313--319.

\bibitem[{Sugiyama, Suzuki, and Kanamori(2012)}]{sugiyama2012density}
Sugiyama, M.; Suzuki, T.; and Kanamori, T. 2012.
\newblock \emph{Density ratio estimation in machine learning}.
\newblock Cambridge University Press.

\bibitem[{Sun et~al.(2023)Sun, Fang, Wu, Wang, and Cao}]{sun2023eva}
Sun, Q.; Fang, Y.; Wu, L.; Wang, X.; and Cao, Y. 2023.
\newblock Eva-clip: Improved training techniques for clip at scale.
\newblock \emph{arXiv preprint arXiv:2303.15389}.

\bibitem[{Vahid et~al.(2023)Vahid, Brown, Steen, Zhang, Jeon, Kang, Gentles,
  and Newman}]{vahid2023high}
Vahid, M.~R.; Brown, E.~L.; Steen, C.~B.; Zhang, W.; Jeon, H.~S.; Kang, M.;
  Gentles, A.~J.; and Newman, A.~M. 2023.
\newblock High-resolution alignment of single-cell and spatial transcriptomes
  with CytoSPACE.
\newblock \emph{Nature biotechnology}, 41(11): 1543--1548.

\bibitem[{Wang and Isola(2020)}]{wang2020understanding}
Wang, T.; and Isola, P. 2020.
\newblock Understanding contrastive representation learning through alignment
  and uniformity on the hypersphere.
\newblock In \emph{International conference on machine learning}, 9929--9939.
  PMLR.

\bibitem[{Wang et~al.(2024{\natexlab{a}})Wang, Huang, Price, and
  Li}]{wang2024cross}
Wang, X.; Huang, X.; Price, S.~J.; and Li, C. 2024{\natexlab{a}}.
\newblock Cross-modal Diffusion Modelling for Super-resolved Spatial
  Transcriptomics.
\newblock \emph{arXiv preprint arXiv:2404.12973}.

\bibitem[{Wang et~al.(2024{\natexlab{b}})Wang, Liu, Huang, Xiong, and
  Zhang}]{wang2024multi}
Wang, Y.; Liu, X.; Huang, F.; Xiong, Z.; and Zhang, W. 2024{\natexlab{b}}.
\newblock A Multi-Modal Contrastive Diffusion Model for Therapeutic Peptide
  Generation.
\newblock In \emph{Proceedings of the AAAI Conference on Artificial
  Intelligence}, volume~38, 3--11.

\bibitem[{Wang et~al.(2023)Wang, Zhao, Huang, Liu, Yin, Tang, Li, Wang, Zhang,
  and Zhao}]{wang2023connecting}
Wang, Z.; Zhao, Y.; Huang, H.; Liu, J.; Yin, A.; Tang, L.; Li, L.; Wang, Y.;
  Zhang, Z.; and Zhao, Z. 2023.
\newblock Connecting multi-modal contrastive representations.
\newblock \emph{Advances in Neural Information Processing Systems}, 36:
  22099--22114.

\bibitem[{Xie et~al.(2024)Xie, Pang, Chung, Perciani, MacParland, Wang, and
  Bader}]{xie2024spatially}
Xie, R.; Pang, K.; Chung, S.; Perciani, C.; MacParland, S.; Wang, B.; and
  Bader, G. 2024.
\newblock Spatially Resolved Gene Expression Prediction from Histology Images
  via Bi-modal Contrastive Learning.
\newblock \emph{Advances in Neural Information Processing Systems}, 36.

\bibitem[{Zhang et~al.(2024)Zhang, Schroeder, Yan, Yang, Hu, Lee, Cho, Susztak,
  Xu, Feldman et~al.}]{zhang2024inferring}
Zhang, D.; Schroeder, A.; Yan, H.; Yang, H.; Hu, J.; Lee, M.~Y.; Cho, K.~S.;
  Susztak, K.; Xu, G.~X.; Feldman, M.~D.; et~al. 2024.
\newblock Inferring super-resolution tissue architecture by integrating spatial
  transcriptomics with histology.
\newblock \emph{Nature Biotechnology}, 1--6.

\bibitem[{Zhang, Rao, and Agrawala(2023)}]{zhang2023adding}
Zhang, L.; Rao, A.; and Agrawala, M. 2023.
\newblock Adding conditional control to text-to-image diffusion models.
\newblock In \emph{Proceedings of the IEEE/CVF International Conference on
  Computer Vision}, 3836--3847.

\bibitem[{Zhao et~al.(2024)Zhao, Chen, Chen, Bao, Hao, Yuan, and
  Wong}]{zhao2024uni}
Zhao, S.; Chen, D.; Chen, Y.-C.; Bao, J.; Hao, S.; Yuan, L.; and Wong, K.-Y.~K.
  2024.
\newblock Uni-controlnet: All-in-one control to text-to-image diffusion models.
\newblock \emph{Advances in Neural Information Processing Systems}, 36.

\bibitem[{Zhou et~al.(2018)Zhou, Rahman~Siddiquee, Tajbakhsh, and
  Liang}]{zhou2018unet++}
Zhou, Z.; Rahman~Siddiquee, M.~M.; Tajbakhsh, N.; and Liang, J. 2018.
\newblock Unet++: A nested u-net architecture for medical image segmentation.
\newblock In \emph{Deep Learning in Medical Image Analysis and Multimodal
  Learning for Clinical Decision Support: 4th International Workshop, DLMIA
  2018, and 8th International Workshop, ML-CDS 2018, Held in Conjunction with
  MICCAI 2018, Granada, Spain, September 20, 2018, Proceedings 4}, 3--11.
  Springer.

\bibitem[{Zhu et~al.(2022)Zhu, Wu, Zhu, Jiang, Tang, Zhang, Liu, and
  Loy}]{zhu2022celebv}
Zhu, H.; Wu, W.; Zhu, W.; Jiang, L.; Tang, S.; Zhang, L.; Liu, Z.; and Loy,
  C.~C. 2022.
\newblock CelebV-HQ: A large-scale video facial attributes dataset.
\newblock In \emph{European conference on computer vision}, 650--667. Springer.

\end{thebibliography}
\clearpage

\end{document}